\documentclass[runningheads]{llncs}
\usepackage{graphicx}
\usepackage{subcaption}
\usepackage{comment}
\usepackage{amsmath,amssymb} 
\usepackage{color}
\usepackage{epsfig}
\usepackage{booktabs}
\usepackage{array}
\usepackage{multirow}
\usepackage{multicol}
\usepackage{siunitx}
\usepackage{lipsum}
\usepackage{hyperref}
\usepackage[font=footnotesize]{caption}
\usepackage[width=122mm,left=12mm,paperwidth=146mm,height=193mm,top=12mm,paperheight=217mm]{geometry}

\newcommand{\parheader}{\smallskip\noindent\textbf}
\newcommand{\STAB}[1]{\begin{tabular}{@{}c@{}}#1\end{tabular}}

\def\code{{\color{magenta}{\url{http://github.com/tonyngjichun/SOLAR}}}}

\DeclareRobustCommand\onedot{\futurelet\@let@token\@onedot}

\def\onedot{~}
\def\eg{\emph{e.g.}\onedot} 
\def\ie{\emph{i.e.}\onedot} 
 
 \def\vs{\emph{vs.}\onedot}
 
\def\etal{\emph{et al.}\onedot}

\begin{document}
\pagestyle{headings}
\mainmatter
\def\ECCVSubNumber{4963}  

\title{
    SOLAR: Second-Order Loss and Attention for Image Retrieval
} 

\titlerunning{SOLAR: Second-Order Loss and Attention for Image Retrieval}
%

\author{Tony Ng\inst{1} \and Vassileios Balntas\inst{2} \and Yurun Tian\inst{1} \and Krystian Mikolajczyk\inst{1}}
\authorrunning{T. Ng et al.}
%
\institute{MatchLab, Imperial College London\\ \and
Facebook Reality Labs \\
\email{\{tony.ng14, y.tian, k.mikolajczyk\}@imperial.ac.uk} \\ 
\email{vassileios@fb.com}
}
\maketitle


\begin{abstract}
   Recent works in deep-learning have shown that second-order information is beneficial in many computer-vision tasks. Second-order information can be enforced both in the spatial context and the abstract feature dimensions. In this work, we explore two second-order components.
   One is focused on second-order spatial information to increase the performance of image descriptors, both local and global.  It is used to re-weight feature maps, and thus emphasise salient image locations that are subsequently used for description. 
   The second component is concerned with a second-order similarity (SOS) loss, that we extend to global descriptors for image retrieval, and is used to enhance the triplet loss with hard-negative mining.
   We validate our approach on two different tasks and datasets for image retrieval and image matching. The results show that our two second-order components complement each other, bringing significant performance improvements in both tasks and lead to state-of-the-art results across the public benchmarks.
   Code available at: \code 
\end{abstract}


\section{Introduction}
\begin{figure}[t!]
    \begin{center}
        \includegraphics[width=1.\linewidth, trim={
        0pt, 15pt, 0pt, 0pt}, clip]{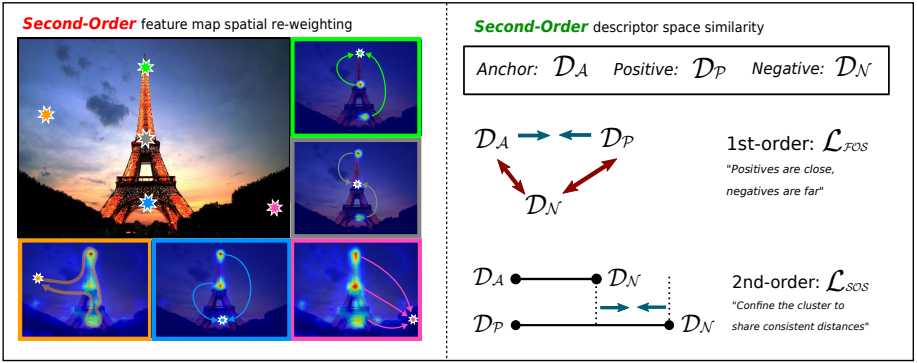}
    \end{center}
    \caption{Illustration of our SOLAR ({\bf S}econd-{\bf O}rder {\bf L}oss and {\bf A}ttention for image {\bf R}etrieval) descriptor. \textbf{Left.} We exploit second-order spatial relations, re-weighting the feature maps to give a better global representation of the image. \textbf{Right.} We also apply second-order similarity of learning discriptor distances during training of SOLAR. 
    }
    \label{fig:teaser}
\end{figure}

Second-order information is receiving increasing attention in computer-vision.
It can be exploited in image retrieval in form of spatial auto-correlation of features, or by second-order similarities in a metric space.
Bilinear features~\cite{Carreira2012Second,CompBiPool,Lin2015Bilinear} compute second-order correlation, but significantly expand feature dimensions,  requiring subsequent dimensionality reduction.
Second-order (self) attention, successful in natural-language processing (NLP)~\cite{transformer}, tackles the dimensionality problem with a multi-headed approach and is hence studied extensively in various vision  areas~\cite{Non-local,Second-Order-ReID,Self-Attn-GANs,ANN}.
Although recent deep-learning based global descriptors provide effective ways to aggregate features into a compact global vector, they have not explored the correlations between features within a feature map. 
Meanwhile, second-order similarity~\cite{YurunNet} has recently been shown to improve patch descriptors for image matching, and has been widely adopted in different vision tasks.
In this work, we exploit the second-order relations between features at different spatial locations and combine with second-order descriptor similarity to improve feature descriptors for image retrieval and matching.
This is illustrated in Fig.~\ref{fig:teaser}. On the left, we learn optimal relative feature contribution spatially (colours of the stars correspond to the frame borders showing the attention for that location). On the right, we use second-order similarity in the descriptor space to make the distance between clusters consistent.

Our main contributions are the following:

\noindent{\bf a)} We combine the second-order spatial attention and the second-order descriptor loss to improve image features for retrieval and matching.\\
\noindent{\bf b)} We show how to combine second-order attention for consecutive feature maps at different resolution to improve the descriptors and we perform a thorough ablation study on its effects.\\
\noindent{\bf c)} We demonstrate that the combination of second-order spatial information and  similarity loss 
generalises well in the context of local and global descriptor learning.\\
\noindent{\bf d)} We validate our method with extensive evaluation on two public benchmarks for image retrieval and matching, showing significant improvements compared to the state-of-the-art.

\section{Related Work}
Methods for image retrieval~\cite{Three-Things-Obj-Retr,HammingEmbedding,OxfordBuildings,Paris,ROxf-RPar} and place recognition~\cite{Dislocation,CityLandmarksMobile,DELF} can be divided into two broad categories: \textit{local aggregation} and \textit{global single-pass}. Most methods prior to deep-learning 
were based on \textit{local aggregation}, \eg Bag-of-Words (BoW)~\cite{VideoGoogle} which aggregates a set of handcrafted, SIFT-like~\cite{SURF,SIFT} local features into a single global vector~\cite{MultiVoc,HammingEmbedding,VLAD,VLAD-SSR,OxfordBuildings,Paris,VideoGoogle,SMK,OrientationCovariant}. 
While many of the \textit{local aggregation} methods carried-over into the deep-learning era~\cite{DELF,DeepFisher,D2R}, the  CNNs~\cite{ResNet,AlexNet,VGG} with highly expressive feature maps~\cite{ImageNet} provided an effective approach for global descriptor encoding. 
Early attempts were mostly hybrid methods, exploring CNN features as direct analogies to local descriptors and aggregating them with similar techniques~\cite{NetVLAD,SPoC,DeepFisher}. Later works showed that CNN feature maps can be embedded into a descriptor with a \textit{single-pass} of a pooling operation~\cite{DeepRetrieval,GeM_old,GeM,MaxPool}, while matching the level of performance from \textit{local aggregation} methods. We group these methods into \textit{global single-pass}.

\parheader{Local Aggregation} methods generally consist of two steps. First, local features are detected and described by hand-crafted operators such as SIFT~\cite{SIFT} and SURF~\cite{SURF}, or CNN-based local descriptors~\cite{SPoC,DELF}. Second, the descriptors are combined into a compact vector. Early works on BoW assigned local descriptors to visual words through various size codebooks~\cite{VideoGoogle}. They were then encoded with matching techniques \eg Hamming Embedding~\cite{HammingEmbedding}, Fisher Kernels~\cite{FV1,FV2} and Selective Match Kernels~\cite{SMK}; or with aggregation techniques \eg k-means~\cite{VocabTree,OxfordBuildings} and VLAD~\cite{VLAD,VLAD-SSR}. 
With the advent of CNN descriptors~\cite{L2Net,YurunNet,LIFT}, learnt features~\cite{SPoC,MutliscaleOrderlessPool,ExploitDeepLocal,DELF} led to substantial improvements in challenging, large-scale retrieval benchmarks~\cite{DELF,ROxf-RPar}. 
Some hybrid methods also learn local-to-global encoding~\cite{NetVLAD,NeuralCodes}. A recent state-of-the-art \textit{local aggregation} system~\cite{D2R} considers features only from regions-of-interest~\cite{FasterRCNN}, filtering out the irrelevant ones such as the sky, background and moving objects.

\parheader{Global Single-Pass} methods, in contrast, do not separate the extraction and aggregation steps. Instead, the global descriptor is generated by a single forward-pass through a CNN. Notice that even though hybrid methods use CNN features as local descriptors followed by \textit{local aggregations}~\cite{NetVLAD,ExploitDeepLocal}, thus generating the global descriptor through a forward-pass of a CNN, we do not consider them to be strictly \textit{global single-pass}, as an individual local representation is still required and aggregated with a handcrafted encoding technique. 
In order to aggregate a feature map from a CNN, either a general~\cite{ImageNet} one or fine-tuned on retrieval-specific datasets~\cite{GeM},  a global pooling operation must be applied. Various \textit{global single-pass} methods differ mostly by the pooling operations, which include Max-pooling \cite{MaxPool}, SPoC \cite{SPoC}, CroW \cite{CroW}, R-MAC~\cite{MaxPool} and GeM~\cite{GeM}. 
GeM pooling has been shown to give excellent results in a recent work that optimises a differentiable approximation of the average-precision metric~\cite{APLoss}.

\parheader{Second-Order Attention} mechanisms proved successful in NLP \cite{transformer}. 
It has since gained popularity in various computer-vision tasks, including video classification \cite{Non-local}, GANs \cite{Self-Attn-GANs}, semantics segmentation~\cite{Non-local,ANN} and person reID~\cite{Second-Order-ReID}. 
However, it has not been employed for visual representation and descriptor learning, in particular for image retrieval and matching tasks. 
On the other hand, \textbf{Second-Order Similarity} has only recently been introduced to representation learning \cite{YurunNet} on local patches by confining the second-order distance in clusters to be similar and distributing them in the area of the unit hypersphere of the descriptor space. Our work is the first to exploit the second-order spatial attention in descriptor learning and to combine it with second-order descriptor loss for learning global image representation for retrieval. 

\section{Method}
In this section, we first present the state-of-the-art Generalised-Mean (GeM) pooling~\cite{GeM} which we then extend with our second-order spatial pooling, followed by second-order similarity loss, whitening and descriptor normalisation. 
\subsection{Preliminaries}
From an input image $I \in {\rm I\!R}^{H,W,3}$  processed through a Fully-Convolutional Network (FCN) denoted by $\theta$, we obtain  a feature map $\textbf{f} = \theta(I) \in {\rm I\!R}^{h,w,d}$ where $h, w$  and $d$ are  height, width  and  feature dimensionality, respectively. 
For $h,w > 1$, Generalised-Mean (GeM) pooling was proposed in~\cite{GeM} as a flexible way to aggregate the feature map into a single descriptor vector $\textbf{D} = \text{GeM}(\textbf{f}, p)$. The GeM pooling with learnable parameter $p$ is defined as
\begin{equation}
    \text{GeM}\left(\textbf{f}, ~p\right) = \left( \frac{1}{N} \sum_{i=0}^N f_i^p \right) ^{\frac{1}{p}}.
    \label{eq:GeM}
\end{equation}
%
\subsection{Second-Order Spatial Pooling}
\label{subsec:2oGeM}
\parheader{Motivation.} There are two main motivations for using spatial second-order attention specifically for image retrieval. First,  $p$ in Equation~\ref{eq:GeM} is able to adjust each local contribution from $\textbf{f}$ to the global descriptor $\textbf{D}$ according the their corresponding feature activation, \ie \textit{absolute} magnitude of a feature vector, which is considered a first-order measurement.
Thus, it assumes the independence of various locations in the map and does not include any \textit{relative} contribution of each spatial feature with respect to the other features.

This is followed closely by the second motivation, where in the case of FCNs such as VGG~\cite{VGG} and ResNet~\cite{ResNet}, each local feature that contributes to the global descriptor $\textbf{D}$ has a limited receptive field covering pixels from the input image. Thus, in Equation~\ref{eq:GeM}, for a specific $f_i$, GeM pooling lacks information on its relation to other features $\{f_k: k \neq i\}$ in $\textbf{f}$.
 
Therefore we propose to generate a map $\textbf{f}^{so}$ with local  features $f^{so}_{i,j}$ that reflect the correlations between all spatial locations from within $\textbf{f}^{so}$,  hence the `second-order'.
Ideally, this will allow the model to learn the optimal \textit{relative} contribution of each spatial feature to the final descriptor $\textbf{D}$.

\parheader{Formulation.} Let each location $(i,j)$ in map \textbf{f} correspond to $(i_I, j_I)$ when projected onto the input image $I$. Assuming a rectangular receptive field $R = [R_x, R_y]$
each vector $f_{i,j} \in \textbf{f}$ is a function of the input pixels $I_{\mathcal{R}}$ included in the receptive field $R$.

\begin{figure*}[t!]
    \begin{center}
        \includegraphics[width=1.\linewidth, trim={
        0pt, 0pt, 0pt, 0pt}, clip]{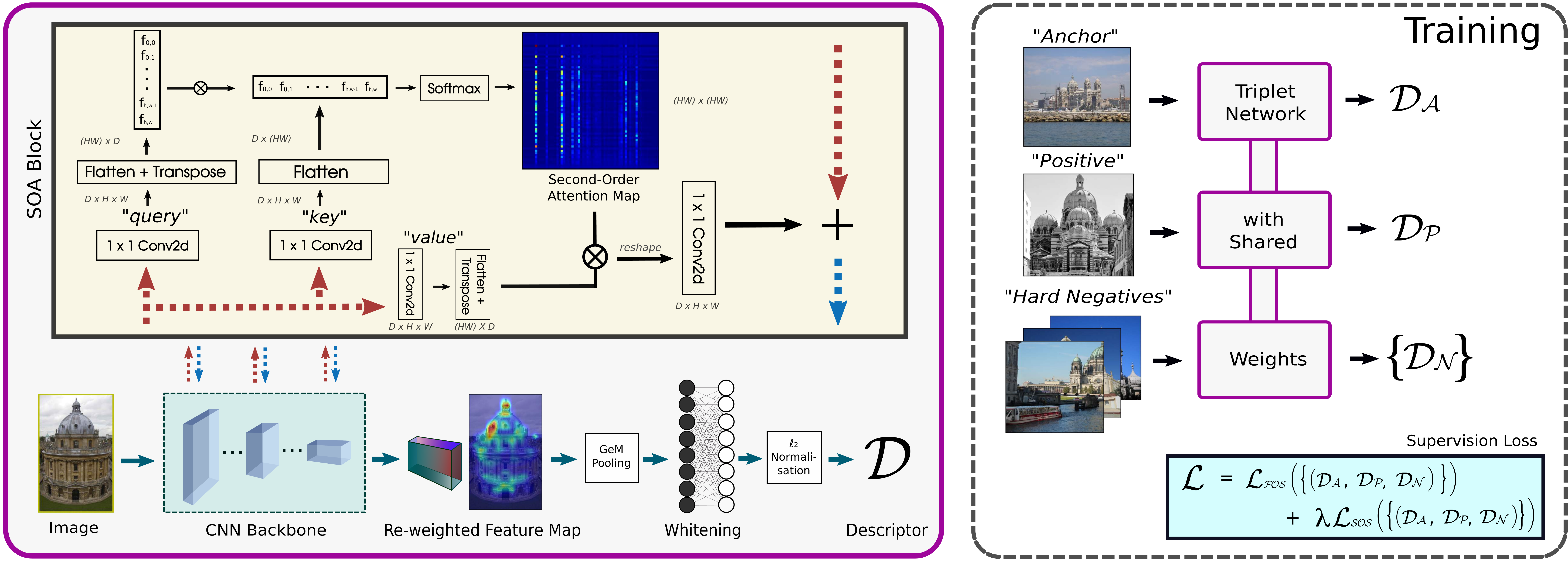}
    \end{center}
    \caption{Pipeline for our proposed global descriptor, SOLAR. We insert a number of {\bf S}econd-{\bf O}rder {\bf A}ttention (SOA) blocks at different levels of a CNN backbone, followed by GeM~\cite{GeM} pooling, whitening and $\ell_2$ normalisation. We train SOLAR using a triplet network  combining first and second-order descriptor loss.}
    \label{fig:pipeline}
\end{figure*}
To incorporate second-order spatial information into the feature pooling, we adopt the non-local block~\cite{Non-local}. 
A visualisation of the concept is shown in the top left of Fig. \ref{fig:pipeline}. 
First, we generate two projections of feature map $\textbf{f}$ termed \textit{query} $\textbf{q}$ head, and \textit{key} $\textbf{k}$ head, each obtained through $1 \times 1$ convolutions\footnote{\label{ftnt:omit}We omit Batch-Norm, ReLU and channel reduction for simplicity. Please refer to our code for the exact model details: \code}. Then, by flattening both tensors, we obtain $\textbf{q}$ and $\textbf{k}$ with shape $d \times hw$. The second-order attention map $\textbf{z}$ is then computed through
\begin{equation}
    \textbf{z} = \text{softmax} (\alpha \cdot \textbf{q}^\intercal \textbf{k}),
    \label{eq:self_attn}
\end{equation}
where $\alpha$ is a scaling factor and $\textbf{z}$ has shape $hw \times hw$, enabling each $f_{i,j}$ to correlate with features from the whole map $\textbf{f}$. 
A third projection of $\textbf{f}$ is then obtained by \textit{value} head $\textbf{v}$, in a similar way to $\textbf{q}$ and $\textbf{k}$, but resulting in shape $hw \times d$.
Finally, 
${\textbf f^{so}}$
map  is obtained from the first-order features $\textbf{f}$ by the second- order attention
\begin{equation}
    \textbf{f$^{so}$} = \textbf{f} + \psi \left( \textbf{z} \times \textbf{v} \right),
    \label{eq:non_local_output}
\end{equation}
where  $\psi$ is another $1 \times 1$ convolution$^{\ref{ftnt:omit}}$ to control the influence of the attention.
Thus, a new feature $f^{so}_{i,j}$ in the second-order map \textbf{f$^{so}$} (reshaped to $h \times w \times d$), is a function of features from all locations in $\textbf{f}$
\begin{equation}
    f^{so}_{i,j} = g(\textbf{z}_{ij} \odot \textbf{f}),
    \label{eq:sec-order}
\end{equation}
where $g$ denotes the combination of all convolutional operations within the non-local block.
We can express  each feature $f^{so}_{i,j}$ as a function of the full input image $f^{so}_{i,j} = \phi \left(i,j,I \right)$, viewed from location $(i,j)$, with $\phi$ as the new FCN with the non-local block(s).
Finally, our extended GeM-pooling 
\begin{equation}
    \text{GeM}\left(\textbf{f$^{so}$}, ~p\right) = \left( \frac{1}{N} \sum_{i=0}^N {f^{so^p}_i} \right) ^{\frac{1}{p}}
    \label{eq:GeMso}
\end{equation}
incorporates second-order information from feature correlations. This is referred to as the \textbf{S}econd-\textbf{O}rder \textbf{A}ttention (SOA) block in the remainder of the paper.
%
\subsection{Second-Order Similarity Loss}
\parheader{First-Order Similarity.}
The triplet loss is a standard formulation for learning first-order descriptors~\cite{balntas2016learning,HardNet,L2Net}. Given a set of triplets formed by anchor, positive and negative images, their corresponding global descriptors are denoted as $\{( \textbf{D}_{a}$, $\textbf{D}_{p}$, $\textbf{D}_{n})\}$. The triplet loss with margin $m$ can be considered as first-order in the descriptor space 
\begin{equation}
    \mathcal{L}_{FOS} = \frac{1}{|\{ (\textbf{D}_{a},\textbf{D}_{p},\textbf{D}_{n})\}|} \sum_{\{ (\textbf{D}_{a},\textbf{D}_{p},\textbf{D}_{n})\}} \text{max} \left(0, \|\textbf{D}_{a} - \textbf{D}_{p}\|^2 - \|\textbf{D}_{a} - \textbf{D}_{n}\|^2 + m  \right)
    \label{eq:triplet}
\end{equation}
\parheader{Second-Order Similarity.}
Following SOSNet~\cite{YurunNet} in local features, a second-order similarity loss can also be applied to global descriptors. We hard-mine negative pairs as in \cite{GeM} and calculate the SOS loss for our descriptors 
\begin{equation}
    \mathcal{L}_{SOS} = \frac{1}{|\{ (\textbf{D}_{a},\textbf{D}_{p},\textbf{D}_{n})\}|} \sqrt{\sum_{\{ (\textbf{D}_{a},\textbf{D}_{p},\textbf{D}_{n})\}}  \left( \|\textbf{D}_{a} - \textbf{D}_{n}\|^2 -  \|\textbf{D}_{p} - \textbf{D}_{n}\|^2\right)^2}.
    \label{eq:SOS}
\end{equation}
The final objective function is a combination of  first and second-order loss for global descriptors obtained with second-order spatial attention balanced by  $\lambda$ 
\begin{equation}
    \mathcal{L} = \mathcal{L}_{FOS} + \lambda \mathcal{L}_{SOS}.
    \label{eq:Losss}
\end{equation}
\subsection{Descriptor Whitening}
Whitening operation is crucial for obtaining well performing descriptors. While the original work in GeM~\cite{GeM} used a linear projection for descriptor whitening~\cite{MikolajczykMatasICCV07}, recent experiments\footnote{\label{fn:gem}\url{http://github.com/filipradenovic/cnnimageretrieval-pytorch}} show superior results from whitening operation learnt end-to-end. We follow this new approach, by inserting a bias-enabled fully-connected layer after GeM pooling with $\ell_2$-norm, and train it end-to-end.
\subsection{Network Architecture and Training}
The pipeline of our proposed method is shown in Fig. \ref{fig:pipeline}. The SOA blocks are insert-able at any feature maps (including intermediate ones), as they serve as learnt feature attention mechanisms. During training all triplets are passed through shared weight networks. Hard-negative mining is also performed at the start of every epoch from a random pool of negatives and it is assured that no negatives from each triplet are from the same scene / landmark class. This is to provide high sample variability from within the mini-batch. Details are described in Section~\ref{subsec:impl}.


\section{Results on Large-Scale Image Retrieval}
\label{sec:results}
In this section, we present results of SOLAR on large-scale image retrieval tasks and compare to the existing methods, both \textit{local aggregation} and \textit{global single-pass}. 

\subsection{Datasets}

\parheader{Google Landmarks 18 (GL18)}~\cite{D2R} is an extension to the original 
Kaggle challenge~\cite{DELF} dataset. It contains over 1.2~million photos from 15$k$ landmarks around the world. These landmarks cover a wide-range of classes from historic cities to modern metropolitan areas to nature scenery. GL18 also contains over 80$k$ bounding boxes singling out the most prominent landmark in each image. In this work it serves as a semi-automatically labelled training dataset. 

\parheader{Revisited Oxford and Paris}~\cite{ROxf-RPar} is the commonly used dataset for evaluating the performance of global descriptors on large-scale image retrieval tasks. 
 Oxford~\cite{OxfordBuildings} and Paris~\cite{Paris} datasets were recently revisited by removing  annotation errors and adding new images. The Revisted-Oxford ($\mathcal{R}$Oxf) and Revisited-Paris ($\mathcal{R}$Par) datasets contain 4,993 and 6,322  images respectively, and each with 70 queries by a bounding box depicting the most prominent landmark in that query. The evaluation protocol is divided into three difficulty levels -- \textit{Easy}, \textit{Medium} and \textit{Hard}. The mean average precision (mAP) and mean precision at rank 10 (mP@10) are usually reported as performance metrics. The supplementary 1M-distractors ($\mathcal{R}$1M) database contains 1-million extra images to test the robustness of descriptors, using the same protocols and metrics as in $\mathcal{R}$Oxf-$\mathcal{R}$Par.

\subsection{Comparison to the State-of-the-Art on Image Retrieval}
\label{sec:comparison}
\parheader{SOTA.} Recent works on large-scale image retrieval~\cite{APLoss,D2R,DAME} select GeM~\cite{GeM} trained on the SfM120k dataset with the contrastive loss as the baseline for \textit{global single-pass} methods. 
However, an update on the GitHub repo by GeM's authors$^{\ref{fn:gem}}$ sets the new state-of-the-art results from GeM trained on the GL18~\cite{D2R} dataset, with the triplet loss as in Equation~\ref{eq:triplet}. 
This setting outperforms the recent method  that proposed the AP-loss~\cite{APLoss} trained on GL18,  when evaluated on $\mathcal{R}$Oxf-$\mathcal{R}$Par~\cite{ROxf-RPar}. Therefore, unlike other recent papers, we select GeM~\cite{GeM} trained on GL18 with the triplet loss as our baseline, and we denote it ResNet101-GeM~[SOTA] in Table~\ref{tab:sota}. We also advocate the use of GL18 training dataset as the new standard protocol for large-scale image retrieval.  The inconsistency of training sets that can be observed across different works makes it difficult to assess what performance gains can be attributed to the proposed methods, rather than the training sets.

\begin{table*}[t!]
\caption{Large-scale image retrieval results of our proposed second-order method against the state-of-the-art on $\mathcal{R}$Oxf-$\mathcal{R}$Par~\cite{ROxf-RPar} and their respective $\mathcal{R}$1M-distractors sets. 
We evaluate against the \textit{Medium} and \textit{Hard} protocols with the mAP and mP@10 metrics. 
For \textit{global single-pass} methods, the first term refers to the backbone CNN. [O] denotes results from  off-the-shelf networks pretrained on Imagenet. 
Our method uses ResNet101 with SOA$\dag$ denoting the best configuration described in Table~\ref{tab:retr_sa}. SOLAR$\dag$ is the full proposed method including the \textbf{S}econd-\textbf{O}rder similarity \textbf{L}oss}
\label{tab:sota}
\medskip
\addtolength{\tabcolsep}{-0.0em}
\renewcommand{\arraystretch}{1.13}
\centering
\resizebox{\columnwidth}{!}{
{\small
    \begin{tabular}{ c | l | r r | r r | r r | r r | r r | r r | r r | r r }
    \toprule
    & \large \multirow{3}{*}{Method}  & \multicolumn{8}{c|}{ \large Medium} & \multicolumn{8}{c}{ \large Hard} \\[+0.15em] \cline{3-18}
     & & \multicolumn{2}{c|}{\large \vphantom{M} \normalsize $\mathcal{R}$Oxf } & \multicolumn{2}{c|}{\normalsize  $\mathcal{R}$Oxf+$\mathcal{R}$1M} & \multicolumn{2}{c|}{\normalsize  $\mathcal{R}$Par} & \multicolumn{2}{c|}{\normalsize  $\mathcal{R}$Par+$\mathcal{R}$1M} & \multicolumn{2}{c|}{\normalsize  $\mathcal{R}$Oxf} & \multicolumn{2}{c|}{\normalsize  $\mathcal{R}$Oxf+$\mathcal{R}$1M} & \multicolumn{2}{c|}{\normalsize  $\mathcal{R}$Par} & \multicolumn{2}{c}{\normalsize  $\mathcal{R}$Par+$\mathcal{R}$1M} \\
         & & mAP & mP@10 & mAP & mP@10 &  mAP & mP@10 &  mAP & mP@10 & mAP & mP@10 &  mAP &  mP@10 &  mAP &  mP@10 &  mAP &  mP@10 \\
    \midrule
    \multirow{4}{*}[.5ex]{\STAB{\rotatebox[origin=c]{90}{Local Agg.}}}
    & HesAff-rSIFT-ASMK$^\star$~\cite{SMK-journal} & \num{60.4} & \num{85.6} & \num{45.0} & \num{76.0} & \num{61.2} & \num{97.9}  & \num{42.0} & \num{95.3} &  \num{36.4} & \num{56.7} & \num{25.7} & \num{42.1} & \num{34.5} & \num{80.6} & \num{16.5} & \num{63.4} \\
    & DELF-ASMK$^\star$~\cite{D2R} & \num{65.7} & \num{87.9} & -- & -- & \num{77.1} & \num{98.7} & -- & -- &  \num{41.0} & \num{57.9} & -- & -- & \num{54.6} & \num{90.9} & -- & -- \\
    & DELF-D2R-R-ASMK$^\star$~\cite{D2R}& \num{69.9} & \num{89.0} & -- & -- & \num{78.7} & \num{99.0} & -- & -- &  \num{45.6} & \num{61.9} &  -- & --  & \num{57.7} & \num{93.0} & -- & -- \\
    & \multicolumn{1}{c|}{--- DELF [GL18]~\cite{D2R}} & \num[math-rm=\mathbf]{73.3} & \num[math-rm=\mathbf]{90.0} & \num[math-rm=\mathbf]{61.0} & \num[math-rm=\mathbf]{84.6}  & \num[math-rm=\mathbf]{80.7} & \num[math-rm=\mathbf]{99.1} & \num[math-rm=\mathbf]{60.2} & \num[math-rm=\mathbf]{97.9} &  \num[math-rm=\mathbf]{47.6} & \num[math-rm=\mathbf]{64.3} & \num[math-rm=\mathbf]{33.6} & \num[math-rm=\mathbf]{53.7} & \num[math-rm=\mathbf]{61.3} & \num[math-rm=\mathbf]{93.4} & \num[math-rm=\mathbf]{29.9} & \num[math-rm=\mathbf]{82.4} \\
    \midrule \midrule
    \multirow{5}{*}[-11ex]{\STAB{\rotatebox[origin=c]{90}{Global Single-Pass}}} 
    & AlexNet-GeM~\cite{GeM} &   \num{43.3} & \num{62.1} & \num{24.2} & \num{42.8} & \num{58.0} & \num{91.6}  & \num{29.9} & \num{84.6} &  \num{17.1} & \num{26.2} & \num{9.4} & \num{11.9}  & \num{29.7} & \num{67.6} & \num{8.4} & \num{39.6} \\
    & VGG16-GeM~\cite{GeM} &  \num{61.9} & \num{82.7}  & \num{42.6} & \num{68.1} & \num{69.3} & \num{97.9}  & \num{45.4} & \num{94.1} &  \num{33.7} & \num{51.0} & \num{19.0} & \num{29.4} & \num{44.3} & \num{83.7} & \num{19.1} & \num{64.9} \\ 
    & ResNet101-R-MAC~\cite{DeepRetrieval} &   \num{60.9} & \num{78.1} & \num{39.3} & \num{62.1} & \num{78.9} & \num{96.9} & \num{54.8} & \num{93.9} &  \num{32.4} & \num{50.0} & \num{12.5} & \num{24.9} & \num{59.4}& \num{86.1} & \num{28.0} & \num{70.0} \\ 
    & ResNet101-SPoC~\cite{SPoC} [O] & \num{39.8} & \num{61.0} & \num{21.5} & \num{40.4} & \num{69.2} & \num{96.7} & \num{41.6} & \num{92.0} &  \num{12.4} & \num{23.8} & \num{2.8} & \num{5.6} & \num{44.7}& \num{78.0} & \num{15.3} & \num{54.4} \\ 
    & ResNet101-CroW~\cite{CroW} & \num{41.4} & \num{58.8} & \num{22.5} & \num{40.5} & \num{62.9} & \num{94.4} & \num{34.1} & \num{87.1} &  \num{13.9} & \num{25.7} & \num{3.0} & \num{6.6} & \num{36.9} & \num{77.9} & \num{10.3} & \num{45.1} \\ 
    & ResNet101-GeM~\cite{GeM} [O] & \num{45.8} & \num{66.2} & \num{25.6} & \num{45.1} & \num{69.7} & \num{97.6}  & \num{46.2}  & \num{94.0} &  \num{18.1} & \num{31.3} & \num{4.7}  & \num{13.4}  & \num{47.0} & \num{84.9} & \num{20.3}  & \num{70.4}  \\
    & ResNet101-GeM~\cite{GeM} & \num{64.7} & \num{84.7} & \num{45.2} & \num{71.7} & \num{77.2} & \num[math-rm=\mathbf]{98.1}  & \num{52.3} & \num[math-rm=\mathbf]{95.3} &  \num{38.5} & \num{53.0} & \num{19.9} & \num{34.9} & \num{56.3} & \num{89.1} & \num{24.7} & \num{73.3} \\
    & ResNet101-GeM+DAME~\cite{DAME} & \num{65.3} & \num{85.0} & \num{44.7} & \num{70.1} & \num{77.1} & \num{98.4}  & \num{50.3} & \num{94.6} &  \num{40.4} & \num{56.3} & \num{22.8} & \num{35.6} & \num{56.0} & \num{88.0} & \num{22.0} & \num{69.0} \\
    & ResNet101-GeM+AP~\cite{APLoss} & \num{67.5} & -- & \num{47.5} & -- & \num{80.1} & --  & \num{52.5} & -- &  \num{42.8} & -- & \num{23.2} & -- & \num{60.5} & -- & \num{25.1} & -- \\
    & ResNet101-GeM~[SOTA]~\cite{GeM} & \num{67.3} & \num{84.7} & \num{49.5} & -- & \num{80.6} & \num{96.7} & \num{57.3} & -- &  \num{44.3} & \num{59.7} & \num{25.7} & -- & \num{61.5} & \num{90.7} & \num{29.8} & -- \\
    [+0.3em]\cline{2-18}\addlinespace[0.3em]
    & \multicolumn{16}{l}{\textbf{Ours}} \\
    [+0.3em]\cline{2-18}\addlinespace[0.3em]
    & ResNet101-GeM+SOS~ & \num{67.6} & \num{84.7} & \num{50.0} & \num{73.1} & \num{80.9} & \num{96.6} & \num{57.6} & \num{94.4} &  \num{44.9} & \num{60.1} & \num{26.2} & \num{42.9} & \num{61.9} & \num{91.0} & \num{30.3} & \num{78.9} \\ 
    & ResNet101+SOA$\dag~$ & \num{68.6} & \num{85.7} & \num{51.3} & \num{74.7} & \num{81.4} & \num{96.6} & \num{58.8} & \num{94.6} &  \num{46.9} & \num{62.7} & \num{28.3} & \num{46.0} & \num{63.7} & \num{91.9} & \num{32.4} & \num{80.9} \\ 
    & ResNet101+SOLAR$\dag~$ & \num[math-rm=\mathbf]{69.9} & \num[math-rm=\mathbf]{86.7} & \num[math-rm=\mathbf]{53.5} & \num[math-rm=\mathbf]{76.7} & \num[math-rm=\mathbf]{81.6} & \num{97.1} & \num[math-rm=\mathbf]{59.2} & \num{94.9}&  \num[math-rm=\mathbf]{47.9} & \num[math-rm=\mathbf]{63.0} & \num[math-rm=\mathbf]{29.9} & \num[math-rm=\mathbf]{48.9} & \num[math-rm=\mathbf]{64.5} & \num[math-rm=\mathbf]{93.0} & \num[math-rm=\mathbf]{33.4} & \num[math-rm=\mathbf]{81.6} \\
    \bottomrule
\end{tabular}
}
}

\end{table*}
\parheader{Comparison}
of SOLAR against other state-of-the-art image retrieval methods on the $\mathcal{R}$Oxf-$\mathcal{R}$Par~\cite{ROxf-RPar}  data is presented in Table~\ref{tab:sota}. By adding SOA blocks, we achieve state-of-the-art mAP and mP@10 performance, and improve by a large margin all other \textit{global single-pass} methods, for both \textit{Medium} and \textit{Hard} protocols.
Adding the Second-Order Loss (denoted by SOLAR$\dag$), the results are further improved by 1\%. SOLAR outperforms mAP of the baseline  in the most challenging \textit{Hard} protocol for $\mathcal{R}$Oxf and $\mathcal{R}$Par by significant 3.6\% and 3.0\% gains respectively, as well as 3.3\% and 2.7\% in mP@10. Our method also outperforms the state-of-the-art \textit{local aggregation} method of DELF-D2R-R-ASMK* in mAP on $\mathcal{R}$Oxf-\textit{Hard} by 0.3\%, $\mathcal{R}$Par-\textit{Medium} by 0.9\% and $\mathcal{R}$Par-\textit{Hard} by 3.2\%. 

For $\mathcal{R}$-1M, SOLAR also achieves the state-of-the-art performance across \textit{global single-pass} methods, outperforming in mAP the SOTA
by $4.0\%$ on $\mathcal{R}$Oxf-\textit{Medium}, $4.2\%$ 
on $\mathcal{R}$Oxf-\textit{Hard}; and by $1.9\%$ on $\mathcal{R}$Par-\textit{Medium}, $3.6\%$ on $\mathcal{R}$Par-\textit{Hard}. 
Compared to ResNet101-GeM+AP~\cite{APLoss} the improvements are even higher ($6.0\%$, $6.7\%$, $6.7\%$ and $8.3\%$).
As for \textit{local aggregation}, SOLAR still achieves comparable results in the $\mathcal{R}$-1M set and even outperforms DELF-D2R-R-ASMK* by $3.5\%$ in mAP for $\mathcal{R}$Par-\textit{Hard}.

\parheader{Speed \& Memory Costs.} It should be noted that the memory requirement for \textit{local aggregation} descriptors is much higher than for \textit{global single-pass} e.g. 27.6GB as reported in DELF-D2R-R-ASMK*~\cite{D2R} \vs 7.7GB for GeM~\cite{GeM} \& SOLAR descriptors in the $\mathcal{R}$1M-distractors set. 
SOLAR also runs with a significantly faster speed compared to DELF-D2R-R-ASMK*, \ie 0.15s processing time per image \vs $>$1.5s on a Titan Xp GPU. 
The SOAs in SOLAR only cause an extra 7.4\% cost in inference time compared to GeM.
For the $\mathcal{R}$-1M distractors set, the extraction time difference is a significant 1.5 days \vs weeks required for DELF-D2R-R-ASMK*. Hence, SOLAR is much more suitable for large-scale retrieval tasks given its scalability when compared to \textit{local aggregation} methods, as well as the performance when compared to \textit{global single-pass} methods.

Moreover, we observe that during training the network converges faster and leads to higher performance on the benchmarks when training \textbf{only} the SOAs and the whitening layer, \ie freezing backbone weights. Not only does this greatly reduce the training time, it also indicates that the SOAs are optimised for \textit{re-weighting} the features, as will be described in the following section.

\begin{figure*}[t!]
    \begin{center}
        \includegraphics[width=1.\linewidth, trim={
        0pt, 0pt, 0pt, 0pt}, clip]{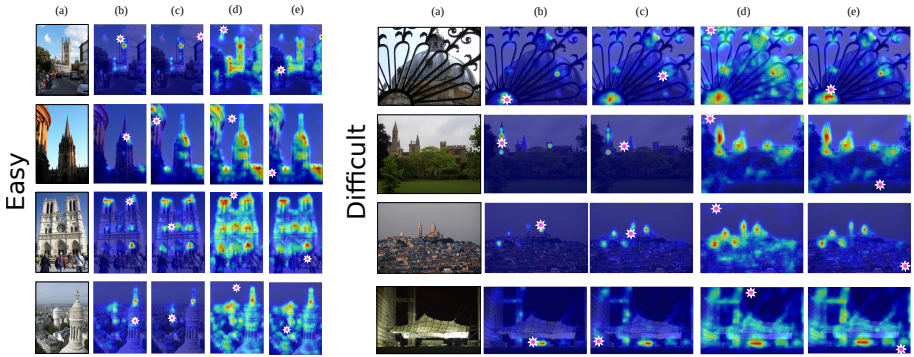}
    \end{center}
    \caption{Qualitative examples of second-order attention maps on the $\mathcal{R}$Oxf-$\mathcal{R}$Par dataset \cite{ROxf-RPar}. 
    Each row depicts (a): the source image and four corresponding second-order attention maps obtained for specific spatial locations (marked by pink stars). For each example, four spatial pixel locations are selected -- (b):  on the dominant landmark, (c): on a secondary landmark, (d): on the sky and (e): on another background part other than the sky. 
    \textbf{Left:} easy examples. \textbf{Right:} difficult examples.}
    \label{fig:image_attn}
\end{figure*}
\subsection{Qualitative Retrieval Results}
\label{subsec:quali_vis}
We visualise the effects of second-order feature map re-weighting in Fig.~\ref{fig:image_attn}. 
For locations in the background ((d) \& (e)), the attention from that feature is sparsely distributed within the  main landmark(s).
On the other hand, when the feature is located within a landmark ((b) \& (c)), the attention is then on highly distinctive regions including informative features from outside of its receptive field. 

This is visible on both, easy examples (\textbf{left} in Fig.~\ref{fig:image_attn}), where there is a clear landmark with distinctive features at similar scales located in the centre and occupies a significant portion of the image, as well as 
challenging examples (\textbf{right} in Fig.~\ref{fig:image_attn}). For example, the top right example has significant occlusion; in the second and third row the landmark is far-away and a large portion of the image is background; and in the bottom row with night-time image. 
We can see that even for these hard examples, the second-order attention maps are consistent.
This provides qualitative evidence that the spatial re-weighting of feature maps, through second-order attentions, is able to assist the network in learning relative contributions from various features into the final descriptor.

\begin{figure*}[t!]
        \includegraphics[width=1\textwidth]{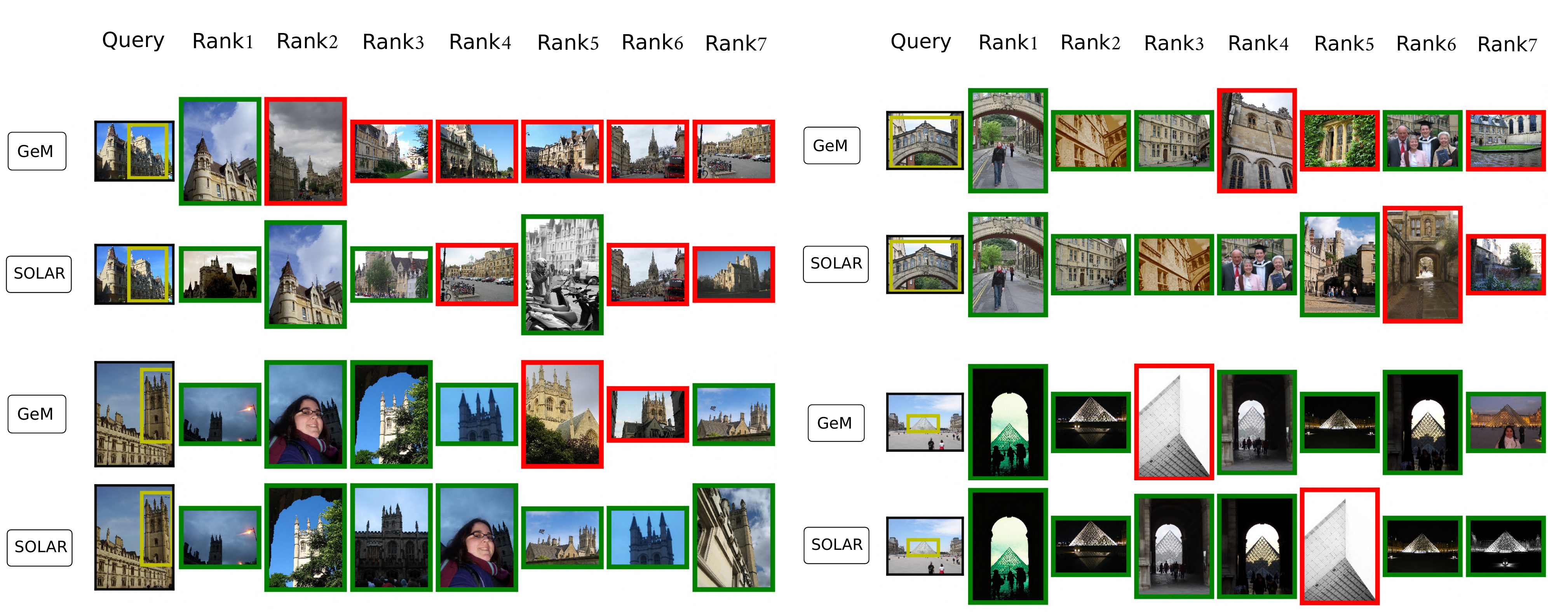}
        \caption{Qualitative comparison between the baseline GeM (top) and SOLAR (bottom).}
        \label{fig:ranking_attention}
\end{figure*}

We also compare the results from image retrieval in Fig.~\ref{fig:ranking_attention} on very challenging examples in $\mathcal{R}$Oxf-\textit{Hard}~\cite{ROxf-RPar}. 
The rows for each example show the query bounding box in yellow,  and the Top-7 ranked retrieved images by the baseline ResNet101+GeM [SOTA]~\cite{GeM} and our ResNet101+SOLAR$\dag$, with green and red borders denoting correct and incorrect retrievals. While GeM performs reasonably well on these examples, it has a tendency to rank high the images containing some similar features, resulting in more false positives.
On the other hand, SOLAR is able to leverage the global correlation from the second-order attentions to increase, in the top few ranks, the number of correct (green) retrievals.

\section{Ablation Study}
\label{sec:ablation}
In this section we evaluate the impact of SOLAR on descriptor performance. We first show how SOLAR leads to learning the optimal feature contribution for pooling a global descriptor from the feature map. Next, we break it down into the two second-order components. 
Lastly, we extend SOLAR to patch datasets to show that it generalises well to local descriptors for image matching task. 
\subsection{Optimal Feature Contribution} 
In Section~\ref{subsec:quali_vis}, we have shown in Fig.~\ref{fig:image_attn}, that  SOAs are effectively \textit{re-weighting} individual feature contributions  into the global descriptor based on their uniqueness within the image. Fig.~\ref{fig:ranking_attention} shows examples of improved retrieval results by SOLAR compared to GeM.
In this section, we conduct a detailed quantitative assessment on the advantages over GeM in optimal feature contributions.

\begin{figure*}[t!]
    \begin{center}
        \includegraphics[width=1\linewidth, trim={
        0pt, 0pt, 0pt, 0pt}, clip]{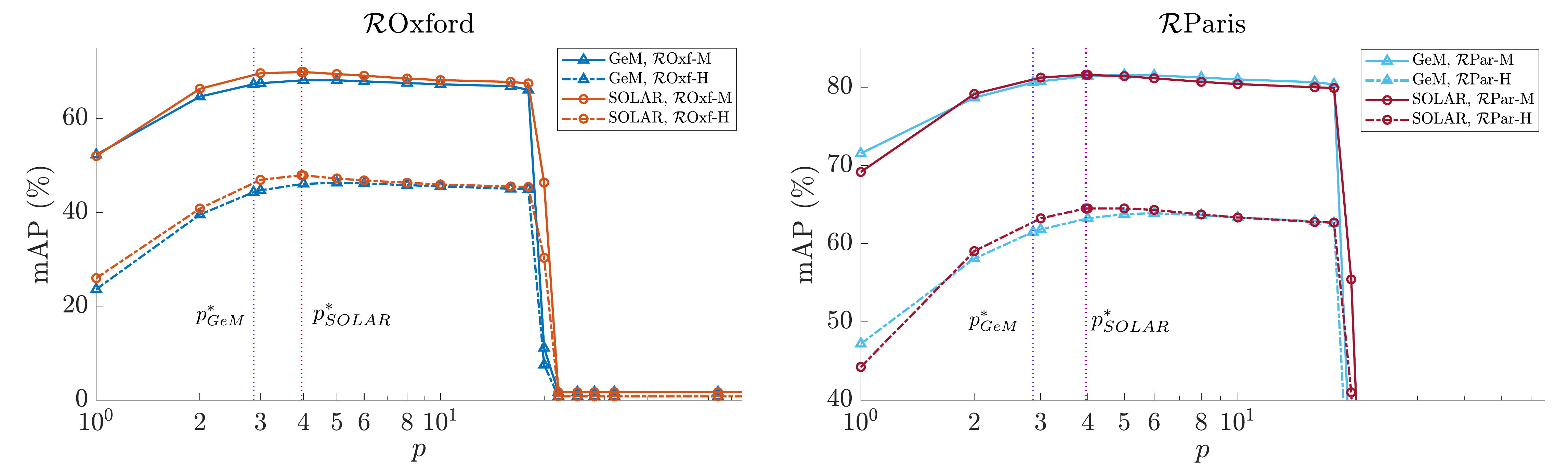}
    \end{center}
    \vspace{-15pt}
    \caption{Comparison of mAP against $p$ on $\mathcal{R}$Oxf-$\mathcal{R}$Par between SOLAR \vs GeM.}
    \label{fig:ablationP}
\end{figure*}

In Fig.~\ref{fig:ablationP} we compare the performance of the baseline (ResNet101-GeM [SOTA]) \vs SOLAR for different values of $p$-norm in Equation~\ref{eq:GeM}. We show the mAP of both methods on the \textit{Hard} and \textit{Medium} protocols of $\mathcal{R}$Oxf-$\mathcal{R}$Par \cite{ROxf-RPar} for $p$ ranging from $p=1$ (\ie equal contribution) to $p=100$ (\ie focused on the strongest features). Note that $p$ is a learnable parameter, we therefore mark the $p$ learnt by each method with dotted-lines on the graphs.
The mAP is clearly increasing as $p$ is raised from $1$ to the learnt value, then drops gradually up to  $p\approx 20$, after which mAP rapidly decreases to a very weak performance.
For high values of $p$, GeM-pooling approaches Max-pooling~\cite{MaxPool}. 
However, $\displaystyle\lim\limits_{p \to \infty} f^p_i = 0 ~ \forall~ |f_i| \leq 1$, causing numerical instabilities in Equation~\ref{eq:GeM}. Hence, in the implementation, feature magnitudes are clipped to a minimum of $10^{-6}$, explaining why mAPs fall after a threshold of $p$ and differ from Max-pooling~\cite{MaxPool}. 

We observe that SOLAR outperforms GeM across most values of $p$, especially in \textit{Hard} examples of both $\mathcal{R}$Oxf and $\mathcal{R}$Par.
More importantly, when comparing the values of $p$ learnt by GeM ($p^*_{GeM}$) and SOLAR ($p^*_{SOLAR}$), $p^*_{SOLAR}$ corresponds to the peak of each of SOLAR's mAP curve, while $p^*_{GeM}$ is sub-optimal to the best mAPs.
This further supports that our SOAs facilitate learning the optimal relative contributions of each feature to the global descriptor.

\subsection{Impact of Second-Order Components on Image Retrieval}
\label{subsec:impact_soa}
The results in Section~\ref{sec:comparison} show that by simultaneously exploiting second-order spatial information through the SOA blocks and second-order descriptor similarity through the SOS loss, we greatly improve  image retrieval performance. 
In this section, we perform an ablation study by gradually incorporating separate second-order components in SOLAR, and discuss the results on image retrieval.

In Table~\ref{tab:retr_sa} we present the impact of adding the second-order loss (SOS) and spatial (SOAs) components, with ResNet101+GeM~[SOTA]~\cite{GeM} as the baseline. 
Firstly, by adding SOS in training, the mAPs improved slightly for $<1\%$.
Then, we look at the effects of adding SOAs into ResNet101~\cite{ResNet}, which contains 5 fully-convolutional blocks \texttt{conv1} to \texttt{conv5\_x}. 
In retrieval, the input image typically has high resolution (1000+ pixels on longer side), inserting SOA blocks before \texttt{conv4\_x} is computationally too expensive 
given the $\mathcal O(n^2)$ complexity of Equation~\ref{eq:self_attn}. 
Table~\ref{tab:retr_sa} shows that our proposed SOA insertions improve retrieval mAP for 0.93\% with SOA$_4$, 1.15\% with SOA$_5$ and 1.78\% with SOA$_{4,5}$. This shows that fine-tuning SOAs alone are more effective than retraining the backbone with SOS.
More importantly, we observe that addition of consecutive SOAs is beneficial and that the improvement brought by fine-tuning on SOA$_5$ is higher than SOA$_4$. 
We believe that this is due to for large images, where the spatial second-order information is still rich and fine-grained even at the last feature map. 
As SOA$_5$ re-weights the last feature map before GeM pooling, it adds second-order spatial information directly into the global descriptor, resulting in a better performance.

Lastly, combining SOS and SOA (\ie SOLAR) gives the best mAPs, and the gain by SOS on SOA ($>1\%$) is more than that of SOS on baseline  ($<1\%$). This further supports that the two second-order components complement each other.

\begin{table*}[t!]
\begin{center}

\caption{
    Ablation study of second-order components on $\mathcal{R}$Oxf-$\mathcal{R}$Par~\cite{ROxf-RPar}. 
    We use ResNet101-GeM~[SOTA]~\cite{GeM} as baseline and incrementally add second-order loss and attention components.
    Results are in mAP for the \textit{Medium} and \textit{Hard} protocols.
    }
    \medskip
{\scriptsize
    \addtolength{\tabcolsep}{0.4em}
    \resizebox{.75\textwidth}{!}{%
    \begin{tabular}{llcccc}
        & & \multicolumn{2}{c}{Medium} & \multicolumn{2}{c}{Hard} \\
        \cmidrule{3-6}
        \multirow{2}{1.8cm}[9pt]{Second-Order Component(s)} & & $\mathcal{R}$Oxf & $\mathcal{R}$Par & $\mathcal{R}$Oxf & $\mathcal{R}$Par \\
        \midrule\midrule
        None (Baseline) & ResNet101-GeM~[SOTA] & 67.3 & 80.6 & 44.3 & 61.5 \\
        \midrule
        Loss (SOS) & ResNet101-GeM+SOS & 67.6 & 80.9 & 44.9 & 61.9 \\
        \midrule
        \multirow{3}{*}[-0pt]{Spatial (SOA)} & ResNet101+SOA$_{4}$ & 68.2 & 81.0 & 45.7 & 62.3 \\
         & ResNet101+SOA$_{5}$ & 68.3 & 81.3 & 45.9 & 62.8\\
         & ResNet101+SOA$_{4,5}$ & 68.6 & 81.4 & 46.9 & 63.7 \\
        \midrule
        Both (SOLAR) & ResNet101+SOLAR & \textbf{69.9} & \textbf{81.6} & \textbf{47.9} & \textbf{64.5} \\
        
        \bottomrule
    \end{tabular}
    }
}
\label{tab:retr_sa}
\end{center}
\end{table*}

\subsection{Generalisation to Image Matching  with Local Descriptors}
\label{subsec:local_resluts}
To validate the generalisation ability of SOLAR besides retrieval with global descriptors, we further test it on local descriptor learning.
Local patches have different statistics than images, containing less semantic information. However, some degree of structure is still present in patches, thus spatial correlation is still informative~\cite{mukundan2019explicit}.
Therefore, we train a local descriptor network with the proposed spatial SOAs. 
With the second-order similarity included in local SOSNet~\cite{YurunNet}, it is straightforward to directly insert SOAs into SOSNet. 

\parheader{Datasets.} In contrast to image retrieval, there are several tasks in different benchmarks to evaluate the performance of local descriptors. Most frequently used are the UBC Patches \cite{ubc2011} and HPatches~\cite{balntas2017hpatches}, as well as other localisation benchmarks that test both feature detectors and descriptors simultaneously. 

\textbf{UBC Patches}~\cite{ubc2011}, consists of three scenes (\textit{liberty}, \textit{notredame}, and \textit{yosemite}) from which corresponding patches are extracted.
Models are trained on one scene and tested on the other two for evaluation. 
Previous works~\cite{balntas2016learning,HardNet,mukundan2019explicit,L2Net,YurunNet} report the false positive rate at 95\% recall (\textbf{FPR@95}) on the 100K test pairs. 
However, the performance on this dataset has saturated, and the limitations of the \textbf{FPR@95} metric have also been pointed out~\cite{hpatchesPAMI}.  Moreover, the evaluation task for UBC is different in nature from retrieval.
Therefore, we leave the results for UBC in the supplementary material and use UBC data only for training, which is a standard protocol for the HPatches benchmark.

\textbf{HPatches}~\cite{balntas2017hpatches} contains over 1.5 million patches extracted from 116 scenes with varying viewpoint and  illumination.
There are three evaluation tasks: \textit{Patch
Verification}, \textit{Image Matching} and \textit{Patch Retrieval}.

\parheader{Impact of SOA at Different Layers.} SOSNet \cite{YurunNet} uses the L2-Net \cite{L2Net} architecture as the backbone. There are 7 convolutional layers in L2-Net which takes a $32\times32$ grayscale input patch and outputs a local descriptor with dimensionality of 128. The L2-Net architecture is presented in the supplementary material.
The SOA block can be inserted at each intermediate feature map except for Layer-7, as the spatial dimension is reduced to $1\times1$ only. 
The earlier the SOA block(s) is inserted, the higher the resolution and  more second-order information can be exploited. However, this  comes at two costs. 
First, the complexity of Equation~\ref{eq:self_attn} is $\mathcal O(n^2)$, where $n$ is the product of the two spatial dimensions. 
Second, the channel depth is shallower at early layers (32 in the first two \vs 128 in the final three layers), \ie each spatial feature in the early layers is less informative.

\begin{figure*}[t!]
\centering
    \includegraphics[width=1\linewidth, trim={
    0pt, 0pt, 0pt, 0pt}, clip]{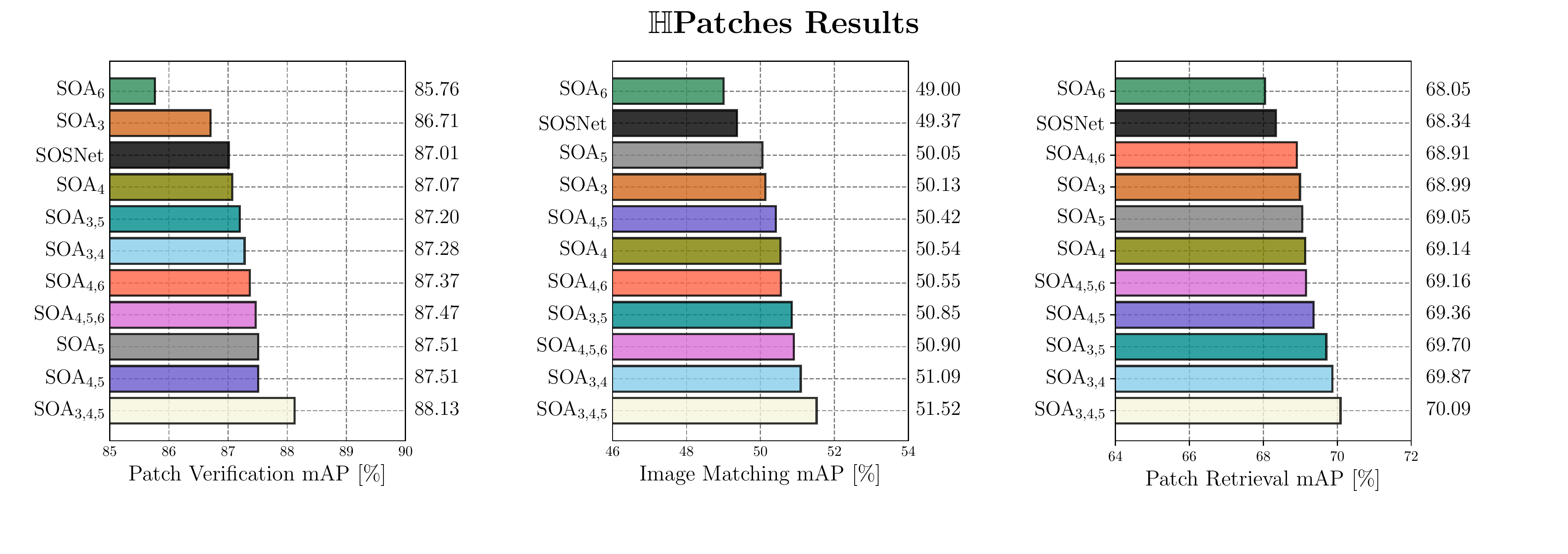}
    \caption{Patch description performance on HPatches. Each of the configurations is denoted as SOA followed by the numbers indicating layers in SOSNet~\cite{YurunNet} backbone after which the blocks are inserted. We train all models with the \textit{liberty} subset of UBC and select the model with the lowest average FPR@95. Patches are resized to $32\times32$.}
    \label{fig:hpatches_results}
\end{figure*}

The results on HPatches with our SOLAR patch descriptors are presented in Fig.~\ref{fig:hpatches_results}.
To investigate how second-order spatial information changes in patch description, 
%
we insert 1 to 3 SOA blocks from between Layers-3 to 7 of L2-Net 
(Layers-1~\& 2 add too much computational cost),
giving the set of results \{SOA${_3}$, SOA${_4}$, SOA${_5}$, SOA$_{6}$, SOA$_{3,4}$, SOA$_{3,5}$, SOA$_{4,5}$, SOA$_{4,6}$, SOA$_{3,4,5}$, SOA$_{4,5,6}$\}.

Models are trained on the \textit{liberty} subset of the UBC dataset~\cite{ubc2011} following standard protocols. We select the best model according to the average {\bf FPR@95} on \textit{notredame} and \textit{yosemite} for each SOA configuration.
Fig.~\ref{fig:hpatches_results} shows that SOAs generally improve \textit{Patch Retrieval} mAP, up to 1.75\% over SOSNet. 
The only exception is SOA$_{6}$ and is due to low spatial resolution of this feature map (only $8\times8$) compared to large images in Section~\ref{subsec:impact_soa}, resulting in less informative second-order spatial correlation. 
This poses a more difficult optimisation task for the SOAs at the final feature levels. 
We notice that SOAs on consecutive levels (SOA$_{3,4}$ $>$ SOA$_{3,5}$ for $0.17\%$, SOA$_{4,5}$ $>$ SOA$_{4,6}$  for $0.45\%$), and across different scales (SOA$_{3,5}$ $>$ SOA$_{4,5}$ for $0.34\%$ despite having fewer parameters) are both beneficial to retrieval, further validating the results from Section~\ref{subsec:impact_soa}.
The results on \textit{Patch Verification} and \textit{Image Matching} are consistent with  \textit{Patch Retrieval}, especially with the ordering \textit{w.r.t.} different SOA configurations. This shows that our SOLAR descriptor also extends well to describing local patches, generalising well between tasks of image retrieval and matching.

\section{Implementation Details}
\label{subsec:impl}

\parheader{GeM+SOLAR.} We start with ResNet101-GeM~\cite{GeM} pre-trained on GL18\footnote{\url{http://cmp.felk.cvut.cz/cnnimageretrieval/data/networks/gl18/}} and fine-tune the SOAs and the whitening layer with Equation~\ref{eq:Losss}.
We train for a maximum of 50 epochs on the same GL18~\cite{D2R} dataset using Adam~\cite{Adam} with an initial learning rate of $1$e$^{-6}$ ($1$e$^{-4}$ for $p$) and exponential decay rate of 0.01. For each epoch 2000 anchors are randomly selected.
The triplets are formed, for every anchor, with 1 positive and 5 hard-negatives mined from 20,000 negative samples, each from a separate landmark, yielding 5 triplets $\{( \textbf{D}_{a}$, $\textbf{D}_{p}$, $\textbf{D}_{n})\}$ for Equations~\ref{eq:triplet} and~\ref{eq:SOS}. The batch-size is 8.
We use margin $m=1.25$ for the triplet loss and $\lambda = 10$ for SOS loss. At test time, we follow \cite{GeM} by passing 3 scales $[1, \sqrt{2}, \tfrac{1}{\sqrt{2}}]$ to the network and taking the average of the output descriptors.

\parheader{SOSNet+SOAs.} We re-implemented SOSNet~\cite{YurunNet} with the details in the original paper to serve as a baseline (100 epochs max). SOAs are inserted and trained with identical settings.
All experiments are implemented in PyTorch~\cite{PyTorch}. For GeM+SOLAR$\dag$, fine-tuning takes roughly 12 hours across 4 1080Ti GPUs. For SOSNET+SOAs, each training takes roughly 5 hours on a single 1080Ti GPU.
\section{Conclusion}
In this work, we propose SOLAR, a global descriptor that utilises second-order information through both spatial attention and descriptor similarity for large-scale image retrieval.
We conduct detailed quantitative and qualitative studies on the impact of incorporating second-order attention that learns to effectively re-weight feature maps, and combine with the second-order information from descriptors similarity to produce better representation for retrieval.
We extend the SOLAR approach to local patch descriptors and show that it improves upon the current state-of-the-art without extra supervision, proving that such second-order combination generalises to different type of data. 
SOLAR achieves state-of-the-art image retrieval performance on the challenging $\mathcal{R}$Paris+1M benchmark compared to similar \textit{global single-pass} methods by a large margin of 3.6\% as well as outperforms \textit{local aggregation} methods by 3.5\%, while running at a fraction of both time and memory costs.
Our approach also improves state-of-the-art for local descriptors in HPatches benchmark by 1.75\%.

\parheader{Acknowledgement.}
This work was supported by UK EPSRC EP/S032398/1 \& EP/N007743/1 grants.
We also thank Giorgos Tolias for providing $\mathcal{R}$-1M results of ResNet101-GeM [SOTA] in Table~\ref{tab:sota}.
%
%
%

%
%
\bibliographystyle{splncs04}
\bibliography{egbib}

\title{
    SOLAR: Second-Order Loss and Attention for Image Retrieval
    \\
    \vspace{1ex}
    \normalsize{Supplementary Material}
}

\author{}
\institute{}
\maketitle
\section{Results Reported in FPR@95 on UBC Patches}
\begin{table*}
\addtolength{\tabcolsep}{0.2em}
\caption{
    \textbf{FPR@95} on the UBC dataset. We compare original SOSNet results~\cite{YurunNet}, our re-implemention with data augmentation -- SOSNet+ (\textit{reimpl.}) and SOSNET+ with the layer numbers after which SOA are inserted. We performed each experiments three times and report the mean and standard deviation. Note that results from SOA$_{4,5,6}$ are not reported as the network did not converge except when trained on the \textit{liberty} subset.}
\vspace{5mm}
\centering
\setlength\extrarowheight{2pt}
{\small
    \resizebox{\textwidth}{!}{%
    \begin{tabular}{lcccccccr}
    \toprule
    Train & &  \multicolumn{2}{c}{Liberty} &  \multicolumn{2}{c}{Notredame} & \multicolumn{2}{c}{Yosemite} &  \\
   \cmidrule(l){1-1} \cmidrule(lr){3-4}\cmidrule(lr){5-6}\cmidrule(lr){7-8}
    Test &  Extra Params & Notredame & Yosemite & Liberty & Yosemite & Liberty & Notredame & Mean\\
    \midrule[\heavyrulewidth]
    SOSNet  & -- & 1.95 & 0.58 & 1.25 & 1.25 & 2.84 & 0.87 & 1.46 \\
    SOSNet+ (\textit{reimpl.}) & -- & 1.31~\scriptsize$\pm$0.01 & 0.46~\scriptsize$\pm$0.04  & 1.21~\scriptsize$\pm$0.06 & 1.07~\scriptsize$\pm$0.03 & 2.25~\scriptsize$\pm$0.03 & 0.80~\scriptsize$\pm$ 0.04 & 1.138~\scriptsize$\pm$0.034 \\
    SOSNet+, SOA$_{3}$ & 12,288 & 1.22~\scriptsize$\pm$ 0.04 & 0.45~\scriptsize$\pm$0.03 & 1.20~\scriptsize$\pm$ 0.03 & \textbf{0.96}~\scriptsize$\pm$ 0.05 & \textbf{2.01}~\scriptsize$\pm$0.10 & 0.73~\scriptsize$\pm$0.04 & \textbf{1.065}~\scriptsize$\pm$0.050 \\
    SOSNet+, SOA$_{4}$  & 12,288 & 1.27~\scriptsize$\pm$0.04  & 0.44~\scriptsize$\pm$0.03 & 1.23~\scriptsize$\pm$0.06 & 0.99~\scriptsize$\pm$0.09 & 2.08~\scriptsize$\pm$0.01 & 0.78~\scriptsize$\pm$0.01 & 1.130~\scriptsize$\pm$0.040 \\
    SOSNet+, SOA$_{5}$  & 22,528 & 1.26~\scriptsize$\pm$0.03 & 0.44~\scriptsize$\pm$0.02 & 1.17~\scriptsize$\pm$0.02 & 0.99~\scriptsize$\pm$0.06 & 2.06~\scriptsize$\pm$0.04 & 0.73~\scriptsize$\pm$0.02 & 1.108~\scriptsize$\pm$0.031 \\
    SOSNet+, SOA$_{6}$ & 22,528 & 1.30~\scriptsize$\pm$0.02  & 0.56~\scriptsize$\pm$0.05 & 1.23~\scriptsize$\pm$0.07 & 1.59~\scriptsize$\pm$0.08 & 2.80~\scriptsize$\pm$0.05 & 0.93~\scriptsize$\pm$0.01 & 1.380~\scriptsize$\pm$0.046 \\
    SOSNet+, SOA$_{3,4}$  & 24,576 & \textbf{1.21}~\scriptsize$\pm$0.02 & \textbf{0.42}~\scriptsize$\pm$0.03 & \textbf{1.15}~\scriptsize$\pm$0.02 & 1.00~\scriptsize$\pm$0.01 & 2.07~\scriptsize$\pm$0.01 & 0.75~\scriptsize$\pm$0.06 & 1.101~\scriptsize$\pm$0.040 \\
    SOSNet+, SOA$_{3,5}$  & 24,576 & 1.27~\scriptsize$\pm$0.06 & 0.45~\scriptsize$\pm$0.05 & 1.30~\scriptsize$\pm$0.02 & \textbf{0.96}~\scriptsize$\pm$0.03 & 2.19~\scriptsize$\pm$0.02 & 0.82~\scriptsize$\pm$0.01 & 1.139~\scriptsize$\pm$0.030 \\
    SOSNet+, SOA$_{4,5}$ & 34,816 & 1.22~\scriptsize$\pm$0.03 & 0.48~\scriptsize$\pm$0.01 & 1.29~\scriptsize$\pm$0.01 & 0.97~\scriptsize$\pm$0.01 & 2.22~\scriptsize$\pm$0.04 & 0.75~\scriptsize$\pm$0.04 &  1.130~\scriptsize$\pm$0.021\\ 
    SOSNet+, SOA$_{4,6}$  & 34,816 & 1.39~\scriptsize$\pm$0.02 & 0.59~\scriptsize$\pm$0.02 & 1.59~\scriptsize$\pm$0.19 & 1.32~\scriptsize$\pm$0.03 & 2.72~\scriptsize$\pm$0.18 & 0.89~\scriptsize$\pm$0.03  & 1.416~\scriptsize$\pm$0.077\\
    SOSNet+, SOA$_{3,4,5}$  & 47,104 & 1.32~\scriptsize$\pm$0.03 & 0.46~\scriptsize$\pm$0.02 & 1.36~\scriptsize$\pm$0.04 & 1.02~\scriptsize$\pm$0.10 & 2.10~\scriptsize$\pm$0.06 & \textbf{0.71}~\scriptsize$\pm$0.02 & 1.147~\scriptsize$\pm$0.047 \\

    \bottomrule
    \end{tabular}
    }
}

\label{tab:ubc}
\end{table*}

The results reported in 
\textbf{FPR@95} on UBC-Patches~\cite{ubc2011} is shown in Table~\ref{tab:ubc}. We present results on each of the six test runs with various configurations of SOA insertions. We did not perform experiments involving SOA$_{1}$ and SOA$_{2}$ as explained in Section 5 in the paper. 
The layers after which SOAs are inserted are based on the L2-Net architecture in Table~\ref{tab:l2net}. We performed experiments on SOA insertion of one to three blocks from between Layers-3 to 7, giving the set of results \{SOA$_{3}$, SOA${_4}$, SOA$_{5}$, SOA$_{6}$, SOA$_{3,4}$, SOA$_{3,5}$, SOA$_{4,5}$, SOA$_{4,6}$, SOA$_{3,4,5}$, SOA$_{4,5,6}$\}. To resolve potential noise, we follow the practice by Mukundan \etal~\cite{mukundan2019explicit} in performing three separate runs for each experiment and reporting the mean value and standard deviation.

Comparing the results of SOSNet with various SOAs inserted in Table~\ref{tab:ubc}, we can see that in general the SOA blocks increase the results slightly with few extra parameters. Agreeing with HPatches results from Fig. 6 in the paper, configurations with SOA$_{6}$ inserted perform noticeably worse when compared to the baseline. We suspect this also due to the same reason of optimisation constraints for low-resolutions at very higher-level feature maps, as discussed in Section 5.3 in the paper. By comparing SOA$_{3,4}$ with SOA$_{3,5}$ and SOA$_{4,5}$ with SOA$_{4,6}$, we observe that SOAs inserted at consecutive feature levels performs noticeably better. One potential explanation would be the immediate sharing of information across consecutive feature maps, allowing for better gradients into the SOA blocks to optimise for feature re-weighting. This also agrees with the improved performance of SOA$_{4,5}$ over single SOA block insertion for ReseNet101 in Section 5.2 of the paper, and HPatches results in the paper.

\vspace{15pt}
\section{L2-Net Architecture}
\vspace{-15pt}
\begin{table}
\vspace{0pt}
\caption{L2-Net~\cite{L2Net} architecture. Note that we only show the convolutional kernel's parameters and intermediate feature map dimension to assist discussion of SOA block insertions. Refer to Tian \etal~\cite{L2Net} for complete details of the architecture including normalisation and activation layers, and different variations of the model.}
\vspace{5mm}
\centering
{\footnotesize
    \addtolength{\tabcolsep}{0.35em}
    \resizebox{.75\textwidth}{!}{%
    \begin{tabular}{rrrrr}
        Layer & Kernel & Stride & Output shape $[h,w,c]$ & Cumulative \# Params.\\
        \midrule\midrule
        1 & $3\times3$ & $1$ & $32,~32,~~~32$ &  288\\
        2 & $3\times3$ & $1$ & $32,~32,~~~32$ & 9,216\\
        3 & $3\times3$ & $2$ & $16,~16,~~~64$ & 18,432\\
        4 & $3\times3$ & $1$ & $16,~16,~~~64$ & 36,864\\
        5 & $3\times3$ & $2$ & $8,~~8,~128$ & 73,728\\
        6 & $3\times3$ & $1$ & $8,~~8,~128$ & 147,756\\
        \midrule
        7 & $8\times8$ & $1$ & $1,~1,~128$ & 285,984\\
    \end{tabular}
    }
}

\label{tab:l2net}
\end{table}

Table~\ref{tab:l2net} shows the L2-Net~\cite{L2Net} architecture, which is used by SOSNet~\cite{YurunNet} and the ablation study from Section 5.3 in the paper. In our implementation of SOSNet and subsequent SOSNet+, SOAs experiments, the patch first passes through an InstanceNorm layer, then each convolution layer is followed by BatchNorm and ReLU (except for after Layer-7 which has no ReLU). Lastly, $\ell_2$-norm is applied to the final 128-dimensional descriptor after Layer-7. During training, dropout of rate 0.1 is added between Layer-6 and Layer-7 to prevent over-fitting.

\section{Second-order attention maps on patches}
\label{subsec:attn_patches}
\begin{figure*}[t!]
    \includegraphics[width=1.\linewidth, trim={
    0pt, 0pt, 0pt, 0pt}, clip]{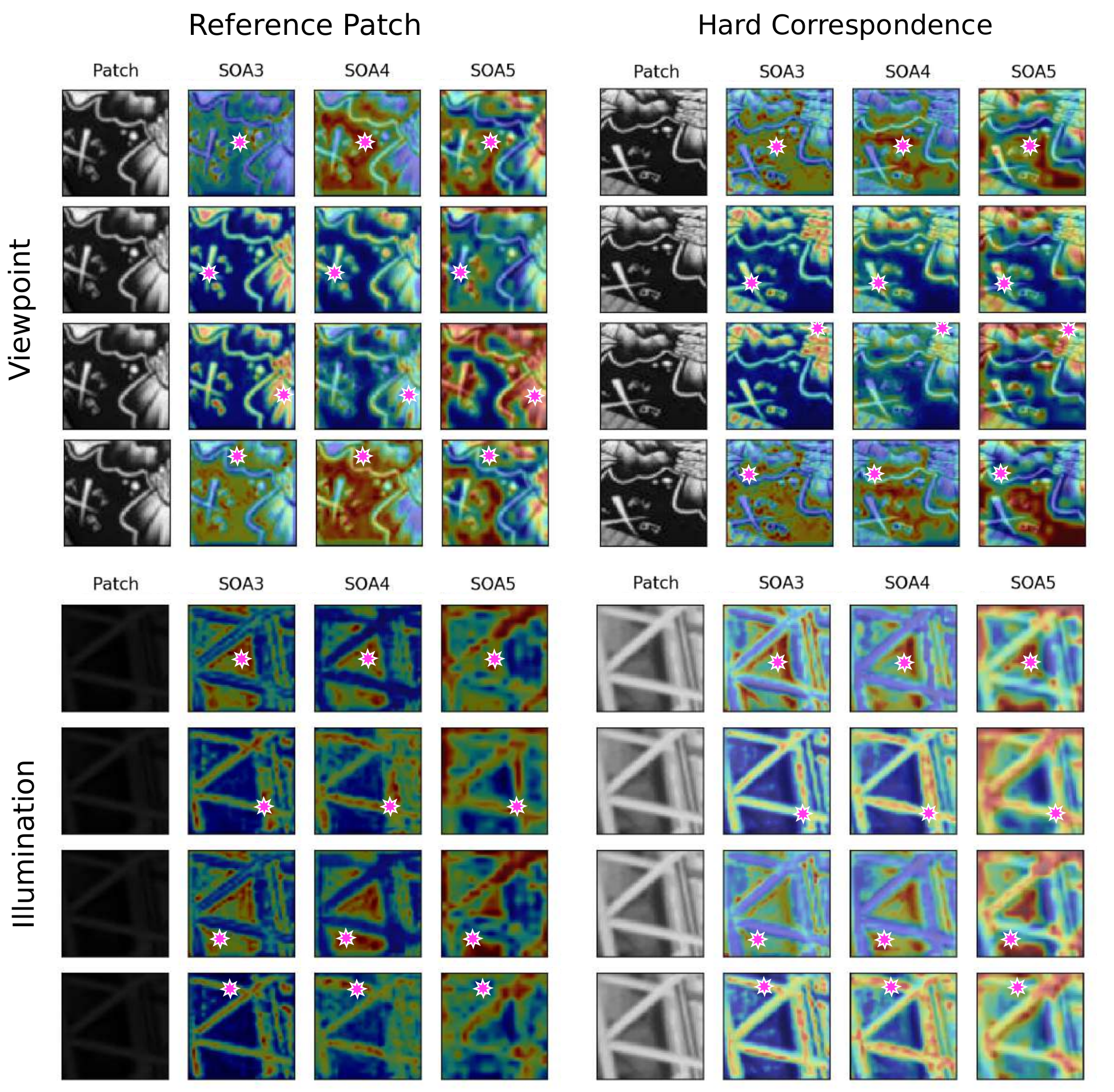}
    \vspace{-5pt}
    \caption{Second-order attention maps for HPatches~\cite{balntas2017hpatches}. \textbf{Left:} reference patch. \textbf{Right:} hard correspondence. \textbf{Top:} viewpoint changes to reference patch. \textbf{Bottom:} illumination changed to reference patch. For each case, we select four pixel locations (pink star) to display the attention maps of SOSNet~\cite{YurunNet}+, SOA$_{3,4,5}$. which has the best results in HPatches evaluation.}
    \vspace{8mm}
\label{fig:local_attn}
\end{figure*}
Fig.~\ref{fig:local_attn} on the next page visualises the second-order attention maps (similar to Figure 4 in the paper) on two example patch correspondences from HPatches~\cite{balntas2017hpatches}. We show two example reference patches and each a `hard' corresponding patch from a sequence with viewpoint (top) and illumination changes (bottom). Firstly we observe that in contrast to large images, the second-order attention at a given spatial location focuses on similar / connected structures within the patch. This is due to much less semantic (and colour) information and lack of distinctive textures in patches compared to large images. Secondly we also observe that the attention maps are invariant to both viewpoint and illumination changes. As we compare the reference patch to the hard correspondence, the attentions between are consistent across all three levels in SOA$_{3,4,5}$.

\end{document}